\documentclass{article}
\usepackage{paper_layout,times}
\usepackage{amsmath,amsfonts,bm}

\def\eqref#1{equation~\ref{#1}}
\def\1{\bm{1}}

\DeclareMathAlphabet{\mathsfit}{\encodingdefault}{\sfdefault}{m}{sl}
\SetMathAlphabet{\mathsfit}{bold}{\encodingdefault}{\sfdefault}{bx}{n}

\usepackage{graphicx}
\usepackage{booktabs}
\usepackage{capt-of}
\usepackage{siunitx}
\usepackage[table]{xcolor}
\definecolor{tablegroup}{RGB}{240,240,240}
\definecolor{tableresult}{RGB}{229,239,252}
\usepackage{tabularx}
\usepackage{wrapfig}
\usepackage{array}
\usepackage{algorithm}
\usepackage{algpseudocode}
\usepackage{placeins}
\usepackage{flafter}
\usepackage{hyperref}
\hypersetup{hidelinks,pdftitle={EvoOPD: Evolving Privileged Information Strategies for Black-Box On-Policy Distillation},pdfauthor={Ximo Zhu, Ruiqi Liu, Yi Han, Zhengan Wu, Gengsheng Li, ZiAn Wang, Zhu Yuantian, Jin Jiang, Wenhao Wang}}
\usepackage{url}
\usepackage{listings}
\usepackage[most]{tcolorbox}
\definecolor{piError}{RGB}{196,144,151}
\definecolor{piErrorBg}{RGB}{255,251,251}
\definecolor{piErrorTitle}{RGB}{249,238,240}
\definecolor{piBlue}{RGB}{116,151,199}
\definecolor{piBlueBg}{RGB}{250,252,255}
\definecolor{piBlueTitle}{RGB}{234,242,252}
\definecolor{piGreen}{RGB}{114,174,152}
\definecolor{piGreenBg}{RGB}{250,254,252}
\definecolor{piGreenTitle}{RGB}{235,247,241}
\tcbset{paperbox/.style={enhanced,breakable,boxrule=0.4pt,arc=2pt,
  left=7pt,right=7pt,top=5pt,bottom=5pt,before skip=6pt,after skip=6pt,
  coltitle=black,fonttitle=\bfseries\small},
  caseerror/.style={paperbox,colback=piErrorBg,colbacktitle=piErrorTitle,colframe=piError},
  casepi/.style={paperbox,colback=piBlueBg,colbacktitle=piBlueTitle,colframe=piBlue},
  caserepair/.style={paperbox,colback=piGreenBg,colbacktitle=piGreenTitle,colframe=piGreen}}
\lstdefinestyle{paperprompt}{basicstyle=\ttfamily\footnotesize,breaklines=true,
  columns=fullflexible,keepspaces=true,showstringspaces=false,frame=single,
  framerule=0.3pt,framesep=6pt,xleftmargin=6pt,xrightmargin=6pt,
  aboveskip=8pt,belowskip=8pt}
\newcommand{\method}{EvoOPD}
\newcommand{\bank}{\mathcal{B}}
\newcommand{\data}{\mathcal{D}}
\date{}
\begin{document}
\begin{center}
{\LARGE\scshape EvoOPD: Evolving Privileged\\
Information Strategies for\\
Black-Box On-Policy Distillation\par}
\vspace{8pt}
{\normalsize Ximo Zhu\textsuperscript{1,*}, Ruiqi Liu\textsuperscript{2,*}, Yi Han\textsuperscript{3}, Zhengan Wu\textsuperscript{4}, Gengsheng Li\textsuperscript{2},\\[4pt]
ZiAn Wang\textsuperscript{5}, Zhu Yuantian\textsuperscript{2}, Jin Jiang\textsuperscript{6}, Wenhao Wang\textsuperscript{7,\textdagger}\par}
\vspace{5pt}
{\small
\textsuperscript{1}University of the Chinese Academy of Sciences\\[2pt]
\textsuperscript{2}Institute of Automation, Chinese Academy of Sciences\\[2pt]
\textsuperscript{3}Southwest University\quad \textsuperscript{4}Southeast University\quad \textsuperscript{5}Jilin University\\[2pt]
\textsuperscript{6}United Arab Emirates University\quad \textsuperscript{7}Vast Intelligence Lab\par}
\vspace{6pt}
{\small \textsuperscript{*}Equal contribution.\qquad \textsuperscript{\textdagger}Corresponding author.\par}
\end{center}
\vspace{3pt}
\begin{tcolorbox}[enhanced,colback=piBlueTitle,colframe=piBlue,boxrule=0.5pt,arc=2pt,
left=10pt,right=10pt,top=8pt,bottom=8pt,before skip=4pt,after skip=12pt]
\centerline{\large\scshape Abstract}\vspace{4pt}
\small
Proprietary language models exhibit strong reasoning and coding capabilities, but their output distributions are inaccessible, making fine-grained distillation methods that rely on teacher logits difficult to apply directly to black-box models. We observe that rich textual feedback from a black-box teacher can be treated as privileged information (PI) and converted, through a PI-conditioned privileged teacher, into dense supervision signals on the student's own trajectories. However, what PI should be acquired from the teacher to most effectively support student learning remains unclear. PI strategies that differ in where they focus, what information they request, and how much detail they provide can have substantially different effects on the current student, and their relative effectiveness can further change as training progresses. Based on this observation, we propose Evolving Privileged Information Strategies for Black-Box On-Policy Distillation (EvoOPD), which treats PI acquisition itself as an optimization problem and maintains a PI Strategy Bank that evolves with the student state. EvoOPD uses the measured repair utility of PI on the current student to adapt the focus, information type, and granularity of teacher feedback, allowing the acquired information to continually match the student's learning needs. We systematically evaluate EvoOPD on mathematical reasoning and code generation with Qwen3-1.7B, 4B, and 8B students. EvoOPD achieves the best overall performance in all six model-task settings, improving over the strongest baseline in each setting by an average of 2.14 percentage points in mathematics and 1.30 points in code. Further analysis shows that the relative effectiveness of PI acquisition strategies changes with student state, making a fixed strategy difficult to sustain as the best choice throughout distillation.
\end{tcolorbox}
\section{Introduction}
\label{sec:introduction}
\begin{figure}[t]
\centering
\includegraphics[width=\linewidth]{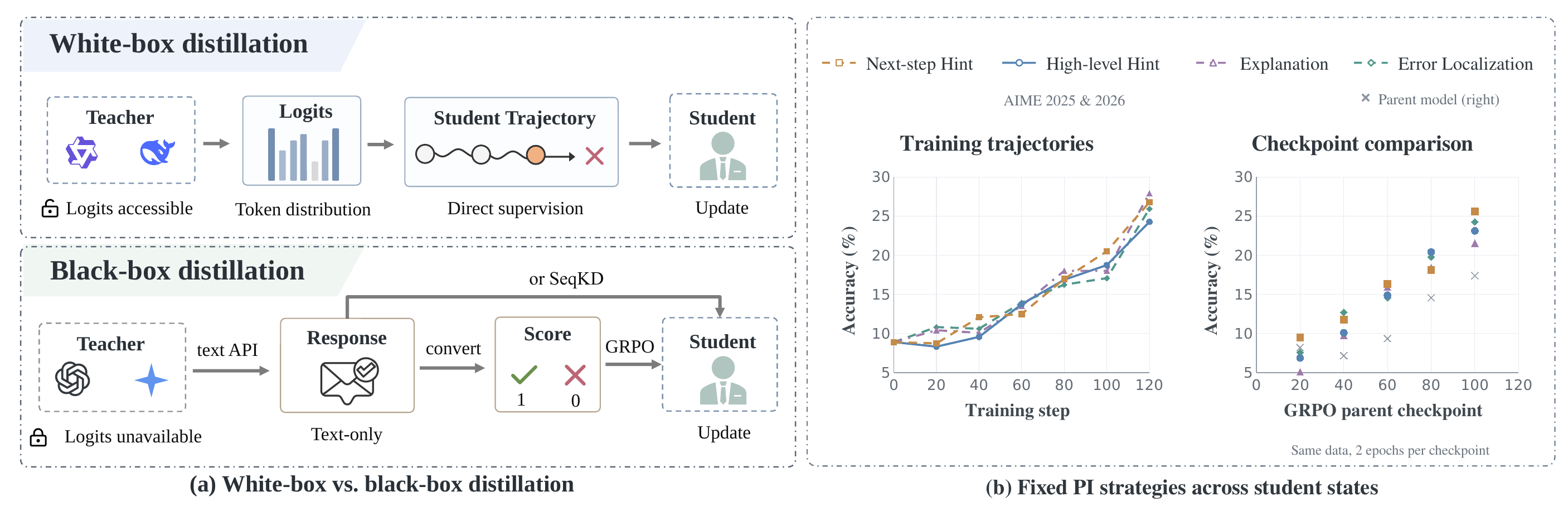}
\caption{\textbf{Motivation for adapting PI acquisition.} (a) White-box distillation uses accessible teacher logits for token-level supervision; black-box distillation uses teacher text for SeqKD or reward-based training. (b) Four fixed PI acquisition strategies evaluated on AIME 2025 and 2026 (16 samples per question). Left: unassisted accuracy along separate training trajectories. Right: unassisted accuracy after training from each GRPO parent checkpoint on the same data for two epochs (20 additional data steps and 80 optimizer updates); gray crosses mark parent-model accuracy before branch training. Strategy rankings change with student state.}
\label{fig:strategies}
\end{figure}

Proprietary language models such as GPT-5 and Gemini exhibit strong reasoning and coding capabilities~\citep{singh2025openai,comanici2025gemini}, motivating research on distilling these capabilities into smaller models. Sequence-level distillation (SeqKD) trains students on teacher-generated answers, but teacher trajectories differ from the student's own generated trajectories~\citep{kim2016sequence,agarwal2024policy}. On-policy distillation (OPD) addresses this mismatch by providing token-level supervision directly on student-generated trajectories, typically using the teacher's output probabilities. As Figure~\ref{fig:strategies}(a) shows, white-box teachers can provide logits directly, whereas black-box teachers such as GPT and Gemini generally expose only text interfaces, leaving their output distributions, hidden states, and parameters inaccessible. Transferring black-box teacher knowledge onto the student's own trajectories through observable text alone is therefore a central challenge in black-box distillation.

Despite lacking access to token probabilities, a black-box teacher can read student responses and provide diagnostic and corrective information as text. We treat this text as privileged information (PI). The student generates fresh unassisted responses and receives binary verifier rewards for group-relative policy optimization. A no-gradient pass of the same student under PI checks whether it assigns lower probability to original tokens in a localized error window. When it does, PI strengthens those tokens' negative advantage; the ordinary policy pass carries the gradient. This uses black-box teacher knowledge without requiring its logits, hidden states, or parameters.

Rich textual feedback is not equally useful for distillation. For the same error, a teacher can localize the mistake, explain its cause, identify a missing constraint, or provide a next-step hint. These forms of feedback differ in focus, information type, and granularity, and may have different learning value for the current student. As the right subplot of Figure~\ref{fig:strategies}(b) shows, different PI acquisition strategies produce different learning outcomes when starting from the same student checkpoint. The utility of teacher feedback therefore depends on the form of information acquired. Beyond how to use teacher feedback, black-box distillation faces a more basic question: what information should be acquired from the black-box teacher? This makes the PI acquisition strategy itself an object of optimization.

These differences between strategies also change during training. As the student improves, both its knowledge gaps and the forms of feedback it can use effectively change. The left subplot of Figure~\ref{fig:strategies}(b) shows that the relative advantages of PI acquisition strategies shift during training: a strategy that is more effective for one student state may lose its advantage for another. PI acquisition strategies therefore need both optimization and continued adjustment to student state.

We propose \emph{Evolving Privileged Information Strategies for Black-Box On-Policy Distillation} (\method). EvoOPD maintains a \emph{PI Strategy Bank} whose rules describe when teacher information is needed, which problems in student responses to focus on, and what type and granularity of feedback to request. The bank is first optimized using measured PI repair utility. During distillation, it guides response-specific PI for current student trajectories; a local likelihood comparison converts that PI into a bounded correction of binary group-relative advantage. As student state changes, EvoOPD periodically reassesses and updates the bank.

We systematically evaluate \method\ on mathematical reasoning and code generation with Qwen3-1.7B, 4B, and 8B. EvoOPD achieves the best Overall performance in all six model--task settings, improving over the strongest baseline in each setting by an average of 2.14 percentage points in mathematics and 1.30 in code. Ablations and mechanism analyses show that structured PI acquisition strategies targeting student errors outperform fixed generic feedback instructions, and that optimizing these strategies further improves unassisted performance. Strategy effectiveness also changes with student state, supporting continued updates during distillation.

\begin{figure}[t!]
\centering
\includegraphics[width=\linewidth]{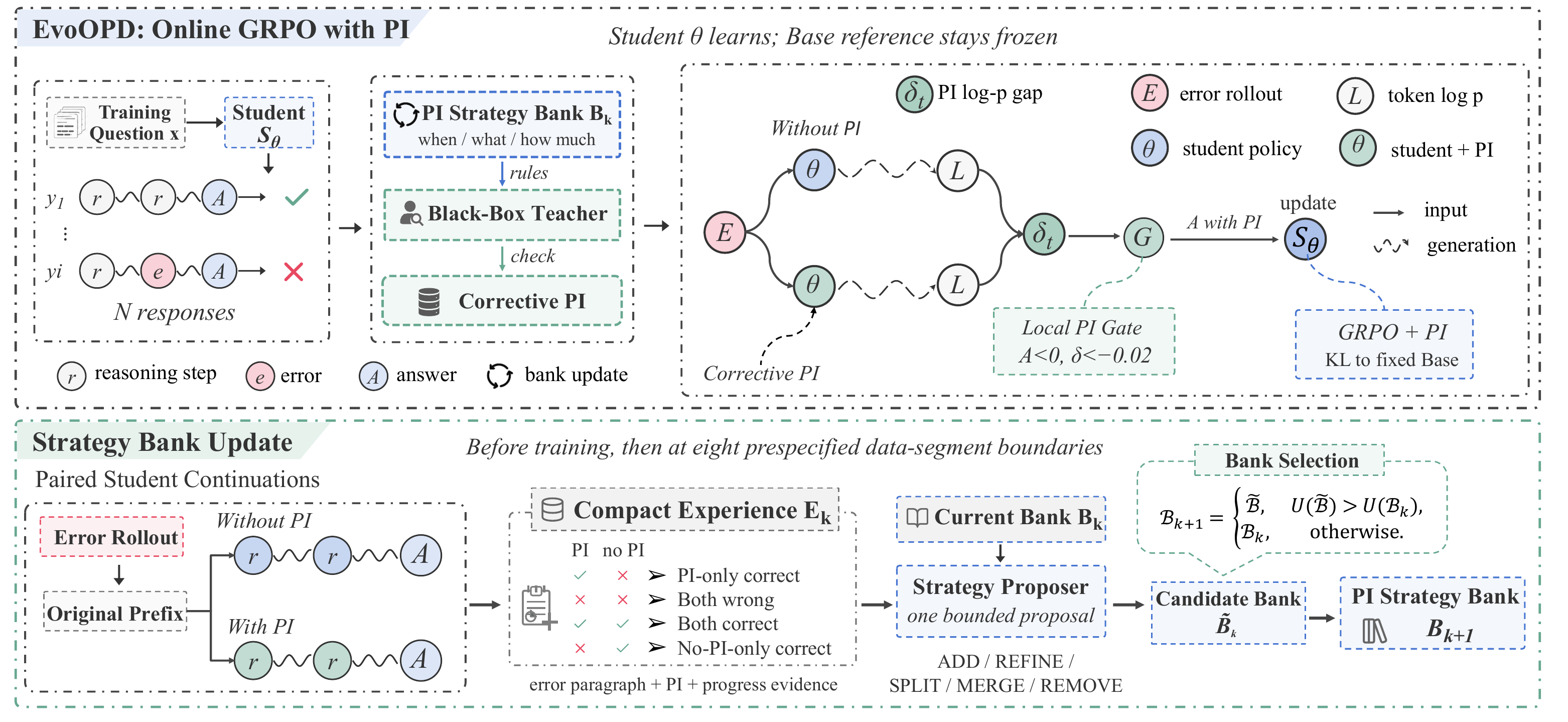}
\caption{\textbf{Overview of \method.} The upper loop acquires response-specific PI under the current strategy bank and uses the student's local response to PI to strengthen negative advantage at supported error tokens. The lower loop adapts PI acquisition by proposing a bounded bank revision and comparing repair utility at the current student on independent selection data.}
\label{fig:overview}
\end{figure}

Our main contributions are:
\begin{itemize}
\setlength{\itemsep}{1pt}
\item We introduce a PI-driven on-policy distillation framework for black-box teachers, which converts textual privileged information into localized dense supervision signals on the student's own trajectories without teacher logits, hidden states, or parameters.
\item We introduce EvoOPD to continually adapt PI acquisition as the student evolves. EvoOPD combines pre-training strategy optimization with continued updates during distillation, using measured PI repair utility at the current student to select PI Strategy Bank revisions and adapt what information is requested from the teacher.
\item We systematically evaluate EvoOPD on mathematical reasoning and code generation with Qwen3-1.7B, 4B, and 8B. EvoOPD improves over the strongest baseline in each setting by an average of 2.14 percentage points in mathematics and 1.30 in code. Ablations and mechanism analyses further reveal that the relative effectiveness of PI acquisition strategies changes with student state.
\end{itemize}

\FloatBarrier
\section{Method}
\label{sec:method}
\subsection{PI Acquisition from a Black-Box Teacher}
For question $x$, the student generates $N$ unassisted responses $y_i$; an independent verifier determines correctness $c_i\in\{0,1\}$. The PI Strategy Bank $\bank_k$ guides the black-box teacher to supply response-specific PI $z_i$ for verified errors, associated with original error windows $W_i$ (Figure~\ref{fig:overview}).

\paragraph{Strategy Bank representation.}
Each rule specifies an applicability condition, a knowledge request, a target policy for selecting the unresolved error, the desired feedback granularity, and guidance for combining complementary rules. The bank is shared across examples, whereas the resulting PI is generated separately for each student response.

Correct responses need no PI. Gold answers and hidden tests remain verifier-only; PI excludes final answers and complete corrected solutions. Appendix~\ref{app:serialization} specifies serialization and localization.

\noindent\textbf{Localization defines the support of PI supervision.}
For an incorrect response, the teacher returns hidden localization metadata and visible PI. Mapping the unresolved error identified by the metadata back to unchanged student tokens defines $W_i$. Only visible PI enters the PI-conditioned forward pass as system guidance; localization metadata is withheld from the student. Teacher feedback thus determines where additional supervision may act without becoming a token-level target. If localization fails or usable PI is unavailable, $W_i=\emptyset$ and training retains the ordinary group-relative update on the original student response under the base training objective.

\subsection{Localized PI Utilization}
Within each question, binary verifier rewards define the unstandardized group-relative advantage:
\begin{equation}
r_i=c_i\in\{0,1\},\qquad
\bar r_x=\frac{1}{N}\sum_{j=1}^{N}r_j,\qquad
A_i=r_i-\bar r_x.
\label{eq:binary_advantage}
\end{equation}
With stop-gradient denoted by $\operatorname{sg}$,
\begin{equation}
\delta_{i,t}=\operatorname{sg}\!\left[
\log S_\theta(y_{i,t}\mid x,z_i,y_{i,<t})-
\log S_\theta(y_{i,t}\mid x,y_{i,<t})
\right].
\label{eq:pi_delta}
\end{equation}
\noindent\textbf{Local response as a counterfactual signal.}
Both scores in Equation~\ref{eq:pi_delta} hold the student parameters, original prefix, and target token fixed; only the visibility of PI changes. The gap therefore measures a local counterfactual response to teacher information, rather than disagreement between models. When $\delta_{i,t}<0$, the student becomes less likely to repeat the original action at that position. This detached probability response does not establish the semantic correctness of PI, and gradients do not pass through the PI-conditioned branch when computing the actor loss.

For margin $\tau>0$, we use
\begin{equation}
\begin{aligned}
g_{i,t}&=\mathbf1[t\in W_i]\,\mathbf1[A_i<0]\,
\mathbf1[\delta_{i,t}<-\tau],\\
\widetilde A_{i,t}&=\operatorname{sg}\!\left[A_i+
\frac{\lambda|A_i|}{\kappa}\,g_{i,t}\,
\operatorname{clip}(\delta_{i,t},-\kappa,\kappa)\right],
\end{aligned}
\label{eq:semantic_gate}
\end{equation}
Here $\lambda>0$ bounds the correction relative to $|A_i|$, and $\kappa>0$ clips the likelihood gap.

\noindent\textbf{Why the correction is one-sided.}
Binary verification and $A_i$ determine whether a trajectory should be promoted or suppressed; PI determines where additional suppression is supported. Textual feedback cannot override correctness or turn an incorrect response into a positive training example. It supplies additional negative evidence only when $t\in W_i$, $A_i<0$, and $\delta_{i,t}<-\tau$ all hold. The bounded correction limits the influence of imperfect feedback. Correct responses, unlocalized tokens, unavailable PI, and probability changes that do not meet the negative-gap threshold retain the base GRPO objective, with no additional PI-induced correction.

Gradients pass through the ordinary clipped policy objective, with a separate KL regularizer to the fixed initial Base model:
\begin{equation}
\mathcal L(\theta;\bank_k)=
\mathcal L_{\mathrm{PPO}}(\theta;\widetilde A)
+\beta\,\mathcal L_{\mathrm{low\_var\_KL}}(S_\theta,S_{\theta_0}).
\label{eq:loss}
\end{equation}
Neither teacher logits, PI tokens, nor corrected answers serve as distillation targets. Appendix~\ref{app:recipe} specifies hyperparameters, policy-ratio clipping, and rollout importance correction.

\subsection{Evolving the PI Acquisition Strategy}
\label{sec:bank}
\label{sec:evolve}
\begin{table}[t!]
\caption{\textbf{Mathematical reasoning after training on DAPO-Math-17K.} Scores are Avg@16 for all three student sizes (\%).}
\label{tab:math}
\centering\small
\setlength{\tabcolsep}{3pt}
\renewcommand{\arraystretch}{1.07}
\begin{tabularx}{\linewidth}{@{}l*{7}{>{\centering\arraybackslash}X}@{}}
\toprule
Method & AIME24 & AIME25 & AIME26 & \shortstack{HMMT25\\Feb} & \shortstack{HMMT25\\Nov} & \shortstack{HMMT26\\Feb} & Overall \\
\midrule
\rowcolor{tablegroup}
\multicolumn{8}{l}{\textit{Qwen3-1.7B}} \\
Base & 12.71 & 10.21 & 7.50 & 5.63 & 3.75 & 6.82 & 7.75 \\
SeqKD & 12.71 & 10.00 & 8.33 & 5.21 & 4.58 & 7.39 & 8.03 \\
OPSD & 10.42 & 7.71 & 6.67 & 5.21 & 3.96 & 7.20 & 6.86 \\
GRPO & 24.79 & 18.96 & 17.08 & 11.88 & 8.75 & 13.45 & 15.78 \\
ROPD & 35.63 & 24.79 & 22.08 & 16.25 & \textbf{16.25} & 21.02 & 22.64 \\
\rowcolor{tableresult}
EvoOPD & \textbf{38.75} & \textbf{27.92} & \textbf{28.13} & \textbf{16.67} & 15.83 & \textbf{22.73} & \textbf{24.97} \\
\addlinespace[2pt]
\rowcolor{tablegroup}
\multicolumn{8}{l}{\textit{Qwen3-4B}} \\
Base & 23.33 & 21.88 & 16.88 & 10.83 & 10.00 & 16.29 & 16.53 \\
SeqKD & 23.13 & 20.42 & 18.13 & 11.04 & 9.58 & 16.67 & 16.50 \\
OPSD & 22.50 & 18.13 & 16.88 & 12.50 & 8.13 & 17.61 & 15.98 \\
GRPO & 49.58 & 42.50 & 40.42 & 25.00 & 28.75 & 28.60 & 35.69 \\
ROPD & 53.33 & \textbf{43.96} & 41.25 & \textbf{25.42} & 35.00 & 30.30 & 38.08 \\
\rowcolor{tableresult}
EvoOPD & \textbf{58.33} & 41.67 & \textbf{47.50} & 24.17 & \textbf{38.33} & \textbf{33.33} & \textbf{40.44} \\
\addlinespace[2pt]
\rowcolor{tablegroup}
\multicolumn{8}{l}{\textit{Qwen3-8B}} \\
Base & 26.04 & 19.38 & 17.08 & 10.63 & 9.79 & 15.72 & 16.43 \\
SeqKD & 25.63 & 21.88 & 16.04 & 9.79 & 10.00 & 16.29 & 16.60 \\
OPSD & 25.83 & 18.96 & 16.67 & 10.83 & 8.75 & 16.48 & 16.26 \\
GRPO & 49.79 & 36.46 & 34.58 & 21.25 & 28.33 & 26.89 & 32.79 \\
ROPD & 61.25 & 51.67 & 51.88 & 31.88 & \textbf{41.25} & 34.09 & 45.15 \\
\rowcolor{tableresult}
EvoOPD & \textbf{63.13} & \textbf{53.13} & \textbf{54.79} & \textbf{33.75} & 40.83 & \textbf{36.74} & \textbf{46.89} \\
\bottomrule
\end{tabularx}
\par\smallskip
\begin{minipage}{\linewidth}\footnotesize\normalfont All evaluations use the student without PI. Overall is weighted by benchmark question count, not a simple mean. Bold indicates the best score within each model size.\end{minipage}
\end{table}

\paragraph{Student-dependent review.}
EvoOPD performs one bank review before gradient training and eight more at prespecified training boundaries. Each review freezes the current student and summarizes recent errors and paired continuations with and without PI into proposal experience $E_k$. The proposer makes one bounded revision, $\widetilde{\bank}=\operatorname{Propose}(\bank_k,E_k)$. On separate selection data $\data_k^{\mathrm{sel}}$, independent of proposal experience, both banks share original responses from this checkpoint but choose their own error locations and PI. The candidate is adopted only when its recovery utility is strictly higher; rejection or a tie leaves the bank unchanged for the next training interval.

\paragraph{Selection by repair utility.}
For all original selection errors $\mathcal E_k$ and binary verifier $V$, recovery utility measures the fraction repaired using a bank's PI:
\begin{equation}
U(B)=\frac{1}{|\mathcal E_k|}\sum_{i\in\mathcal E_k}
\mathbf 1[\text{usable PI under }B]\,
V(x_i,\hat y_{i,B}).
\label{eq:selection_utility}
\end{equation}
Errors without usable PI remain in the denominator. Bank selection follows
\begin{equation}
\bank_{k+1}=
\begin{cases}
\widetilde{\bank},& U(\widetilde{\bank})>U(\bank_k),\\
\bank_k,&\text{otherwise}.
\end{cases}
\label{eq:refine}
\end{equation}
The selected bank guides fresh responses between reviews (Appendices~\ref{app:procedure} and~\ref{app:recipe}).

\paragraph{Why repair utility?}
The Strategy Bank is intended to control what information is acquired from the black-box teacher, rather than to directly optimize the student parameters. We therefore evaluate a candidate bank by whether its PI can help the current student recover from its own errors under a fixed checkpoint. This provides a student-dependent signal for comparing acquisition strategies without requiring teacher logits or differentiating through the black-box teacher. Importantly, repair utility is used only for bank selection; student learning still follows the on-policy objective in Section~2.2. As the student changes, the same acquisition strategy may induce different repair outcomes, motivating repeated evaluation rather than a single offline optimization.

\begin{table}[t!]
\caption{\textbf{Code generation on the OpenThoughts-114k code subset.} Pass@1 is estimated from 16 samples per question for every benchmark and model size (\%).}
\label{tab:code}
\centering\small
\setlength{\tabcolsep}{1.5pt}
\renewcommand{\arraystretch}{1.07}
\begin{tabularx}{\linewidth}{@{}l*{6}{>{\centering\arraybackslash}X}@{}}
\toprule
Method & HumanEval+ & MBPP+ & \shortstack{BigCodeBench\\Hard} & \shortstack{LiveCodeBench\\v5} & \shortstack{LiveCodeBench\\v6} & Overall \\
\midrule
\rowcolor{tablegroup}
\multicolumn{7}{l}{\textit{Qwen3-1.7B}} \\
Base & 63.91 & 49.82 & 7.85 & 11.64 & 15.39 & 34.02 \\
SeqKD & 64.10 & 49.70 & 7.39 & 11.83 & 15.25 & 33.95 \\
OPSD & 62.50 & 50.02 & 8.66 & 11.08 & 15.18 & 33.86 \\
GRPO & 66.88 & \textbf{51.32} & 8.99 & 15.34 & 18.11 & 36.27 \\
ROPD & 64.33 & 48.51 & 9.84 & 13.62 & 17.68 & 34.61 \\
\rowcolor{tableresult}
EvoOPD & \textbf{68.10} & 50.98 & \textbf{11.19} & \textbf{17.55} & \textbf{20.21} & \textbf{37.37} \\
\addlinespace[2pt]
\rowcolor{tablegroup}
\multicolumn{7}{l}{\textit{Qwen3-4B}} \\
Base & 75.69 & 61.14 & 17.82 & 19.12 & 23.29 & 44.02 \\
SeqKD & 75.46 & 61.01 & 17.78 & 18.75 & 23.46 & 43.90 \\
OPSD & 75.30 & 60.96 & 18.20 & 19.05 & 23.18 & 43.92 \\
GRPO & 80.30 & 62.68 & \textbf{18.37} & 21.89 & 24.32 & 46.02 \\
ROPD & 78.28 & 62.05 & 18.33 & 22.42 & 24.86 & 45.64 \\
\rowcolor{tableresult}
EvoOPD & \textbf{81.36} & \textbf{64.07} & 18.24 & \textbf{24.44} & \textbf{26.61} & \textbf{47.48} \\
\addlinespace[2pt]
\rowcolor{tablegroup}
\multicolumn{7}{l}{\textit{Qwen3-8B}} \\
Base & 80.91 & 67.34 & 21.58 & 24.10 & 22.54 & 48.34 \\
SeqKD & 80.56 & 67.15 & 22.30 & 23.54 & 22.68 & 48.25 \\
OPSD & 80.45 & 67.51 & 21.45 & 26.05 & 22.89 & 48.69 \\
GRPO & 82.66 & \textbf{70.19} & 20.73 & 31.85 & 29.29 & 51.94 \\
ROPD & 80.53 & 66.50 & 21.54 & 25.60 & 27.54 & 49.06 \\
\rowcolor{tableresult}
EvoOPD & \textbf{84.07} & 69.94 & \textbf{23.61} & \textbf{34.32} & \textbf{31.68} & \textbf{53.29} \\
\bottomrule
\end{tabularx}
\par\smallskip
\begin{minipage}{\linewidth}\footnotesize\normalfont All evaluations use the student without PI. Overall is weighted by benchmark question count, not a simple mean. Bold indicates the best score within each model size. BigCodeBench uses Hard v0.2.5.\end{minipage}
\end{table}

\section{Experiments}
\label{sec:experiments}

\subsection{Experimental Setup}
Students are Qwen3-1.7B, Qwen3-4B, and Qwen3-8B~\citep{yang2025qwen3}; Gemini-3.6-Flash supplies PI and bank proposals. Training uses DAPO-Math-17K~\citep{yu2026dapo} and the OpenThoughts-114k code subset~\citep{guha2025openthoughtsdatarecipesreasoning}. Evaluation covers AIME and HMMT for mathematics, and EvalPlus, LiveCodeBench versions 5 and 6, and BigCodeBench Hard v0.2.5 for code~\citep{liu2023your,jain2025livecodebench,zhuo2025bigcodebench}.

We compare the original \textbf{Base} model with \textbf{SeqKD}, \textbf{OPSD}, \textbf{GRPO}, and \textbf{ROPD}~\citep{kim2016sequence,zhao2026self,shao2024deepseekmath,fang2026rubric}. All final evaluations use non-thinking contexts without PI or the teacher. Mathematics reports mean sampled-answer correctness and code reports pass@1; Overall weights benchmark question counts. Validation selects final checkpoints. Appendix~\ref{app:details} gives data partitions, method-specific recipes, and decoding settings.

\subsection{Main Results}
\paragraph{Mathematical reasoning.}
ROPD is the strongest Overall baseline at all three model sizes, while EvoOPD further improves Overall at every scale and leads 14 of the 18 model--benchmark comparisons (Table~\ref{tab:math}). The gains are therefore broad across student capacities, although not uniform across individual benchmarks. Since each mathematics benchmark contains only 30--33 questions, the few benchmark-level regressions do not establish a specific reasoning deficit.

Standard OPSD does not improve Overall over Base at any scale in our non-thinking mathematics setting. In one inspected training run, early validation gains also faded later in training, consistent with recent evidence that OPSD can be sensitive to privileged-context construction and training recipe~\citep{zhu2026many,yang2026self,li2026uopsd}. We therefore treat this as setting-specific sensitivity rather than a general failure of self-distillation.

\begin{table}[t!]
\begin{minipage}[t]{0.49\linewidth}
\vspace{0pt}
\setlength{\abovecaptionskip}{0pt}
\caption{\textbf{Qwen3-1.7B ablation.} Math: weighted Overall; code: LCB v6 pass@1 (\%).}
\label{tab:controls}
\centering\small
\setlength{\tabcolsep}{3pt}
\renewcommand{\arraystretch}{1.15}
\begin{tabularx}{\linewidth}{@{}>{\raggedright\arraybackslash}Xcccc@{}}
\toprule
\rowcolor{tablegroup}
\textbf{Condition} & \textbf{Pre} & \textbf{Cont.} & \shortstack{\textbf{Math}\\\textbf{Overall}} & \shortstack{\textbf{LCB v6}\\\textbf{pass@1}} \\
\midrule
No PI & $\times$ & $\times$ & 16.46 & 15.18 \\
Bank-free & $\times$ & $\times$ & 18.00 & 15.81 \\
Fixed Strategy Bank & $\times$ & $\times$ & 20.87 & 17.84 \\
Localization-only & $\checkmark$ & $\checkmark$ & 21.35 & 17.68 \\
Pre-optimized Bank & $\checkmark$ & $\times$ & 22.94 & 18.92 \\
Offline-9 & $\checkmark$ & $\times$ & 23.81 & 18.52 \\
\addlinespace[1pt]
\rowcolor{tableresult}
\method & $\checkmark$ & $\checkmark$ & \textbf{24.97} & \textbf{20.21} \\
\bottomrule
\end{tabularx}
\end{minipage}\hfill
\begin{minipage}[t]{0.49\linewidth}
\vspace{0pt}
\setlength{\abovecaptionskip}{0pt}
\caption{\textbf{Black-box teacher usage for the mathematics ablations.} Input, output, thinking, and total usage are reported in millions of tokens.}
\label{tab:api_budget}
\centering\small
\setlength{\tabcolsep}{2pt}
\renewcommand{\arraystretch}{1.15}
\begin{tabularx}{\linewidth}{@{}>{\raggedright\arraybackslash}X*{4}{S[table-format=3.2,mode=text,detect-weight=true]}@{}}
\toprule
\rowcolor{tablegroup}
\textbf{Condition} & {\textbf{Input}} & {\textbf{Output}} & {\textbf{Thinking}} & {\textbf{Total}} \\
\midrule
No PI & 0.00 & 0.00 & 0.00 & 0.00 \\
Bank-free & 66.96 & 12.33 & 64.81 & 144.10 \\
Pre-optimized Bank & 77.10 & 11.17 & 71.46 & 159.73 \\
Localization-only & 88.54 & 10.45 & 79.14 & 178.13 \\
Offline-9 & 87.91 & 13.39 & 82.91 & 184.21 \\
\addlinespace[1pt]
\rowcolor{tableresult}
\textbf{EvoOPD} & \bfseries 94.88 & \bfseries 9.11 & \bfseries 83.06 & \bfseries 187.05 \\
\bottomrule
\end{tabularx}

\end{minipage}
\end{table}

\noindent\textbf{Code generation.}
Table~\ref{tab:code} shows a different competitive pattern from mathematics. GRPO is the strongest Overall baseline across all three model sizes, while EvoOPD consistently improves further, reaching the best Overall performance at 1.7B, 4B, and 8B. The improvement is visible across HumanEval+ and both LiveCodeBench versions, with MBPP+ and BigCodeBench Hard showing several benchmark-specific exceptions. Overall, the code gains are more moderate than those observed in mathematics. Modest improvements among OPD variants on code benchmarks are also reported by \citet{yang2026influence}, alongside larger gains in reasoning pass@$k$; the size of these effects depends on the benchmark and metric used in the comparison.

The result is nevertheless consistent across model scales: the strongest competing baseline changes from ROPD in mathematics to GRPO in code, while EvoOPD improves the final unassisted student in both domains. Tables~\ref{tab:math} and~\ref{tab:code} compare complete training recipes; Section~\ref{sec:ablation} next examines which parts of PI acquisition and adaptation contribute to these gains.

\subsection{Ablating PI Acquisition and Strategy Updates}
\label{sec:ablation}
Table~\ref{tab:controls} studies three components of EvoOPD: whether teacher feedback is acquired through a reusable strategy, whether PI is converted into local supervision through the student's probability response, and whether the acquisition strategy continues to adapt as the student changes. All conditions use the same gradient-data exposure. No PI retains EvoOPD's binary base objective and optimization recipe, providing the reference point for the added PI components.

\paragraph{Structured PI acquisition.}
Bank-free feedback improves over No PI, showing that response-specific teacher feedback is already useful. Introducing a fixed Strategy Bank yields a larger improvement in both mathematics and code, indicating that reusable rules for when, where, and what to request provide a stronger acquisition interface than unconstrained feedback. Pre-optimizing the bank improves performance further. Together, these comparisons show that not only the feedback itself, but also the strategy used to acquire it, is an optimizable part of black-box distillation.

\paragraph{PI-conditioned scoring beyond localization.}
Localization-only uses the same PI to identify error windows, but replaces EvoOPD's PI-conditioned likelihood signal with a fixed negative correction. Its coefficient is calibrated on independent data so that the total correction magnitude matches that of EvoOPD. EvoOPD instead filters tokens by the PI-induced likelihood change and scales the additional negative advantage accordingly. The full method outperforms Localization-only in both tasks, showing that the student's local probability response to PI provides useful supervision beyond identifying where an error occurs. This comparison evaluates the combined effect of likelihood-based filtering and weighting (Appendix~\ref{app:localization_control}).

\paragraph{Continued adaptation to student state.}
Pre-optimized Bank performs one review before training and then keeps the bank fixed. Offline-9 performs nine reviews, all against the initial student. EvoOPD has the same nine review opportunities, but distributes them across one initial review and eight reviews during training. Its stronger results in both tasks support continued adaptation to student state, beyond increasing the number of reviews at initialization. A fixed Strategy Bank still generates response-specific PI for each new trajectory: the acquisition rule remains fixed, not the PI text. Continued review adapts this rule as the student's needs change.

\paragraph{Teacher usage.}
Table~\ref{tab:api_budget} summarizes recorded teacher-token usage for the mathematics ablations. Offline-9 and EvoOPD both use nine bank-review opportunities and similar totals (184.21M versus 187.05M), but distribute their reviews differently. Together with Table~\ref{tab:controls}, this supports distributing strategy updates across student training rather than concentrating the same review opportunities at initialization. Appendix~\ref{app:api_budget} details the accounting protocol for each reported condition.

\begin{figure*}[t!]
\centering
\includegraphics[width=\linewidth,trim=0 48bp 0 29bp,clip]{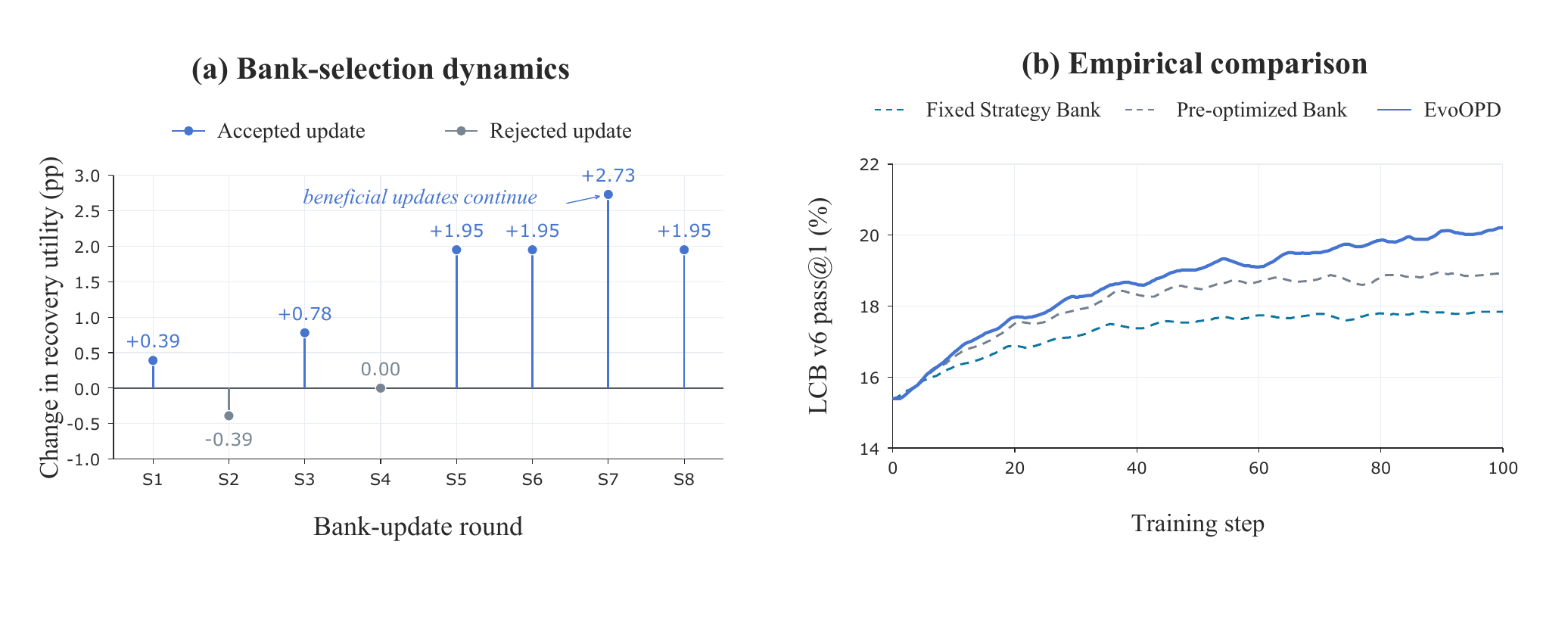}
\caption{\textbf{PI acquisition strategy optimization during distillation.} (a) Candidate-minus-current recovery utility, $\Delta U$, over eight reviews. Positive differences are accepted (blue); otherwise the current bank is retained (gray). (b) LiveCodeBench v6 pass@1 (\%) for three banks, all starting at 15.39\% from the same initial student checkpoint.}
\label{fig:continued}
\end{figure*}

\paragraph{Bank selection during training.}
Figure~\ref{fig:continued}(a) shows that review does not imply replacement: candidate banks are accepted only when they improve recovery utility, while weaker or tied proposals leave the active bank unchanged. Accepted revisions continue to appear in later training stages, indicating that useful acquisition changes remain available after initialization. Figure~\ref{fig:continued}(b) further shows that fixed, pre-optimized, and continually reviewed banks follow increasingly different training trajectories from the same student start, with continued adaptation yielding the strongest final unassisted performance among the three acquisition strategies.

\paragraph{What changes during evolution.}
The accepted revisions change the information requested from the teacher, not merely the wording of the rules. In mathematics, an initial constraint cue that asks for a missing condition is refined into actionable local guidance, and later expands to explicit bounds, structural adjustments, representation recovery, and verification cues. The code bank similarly evolves toward preserving relational structure and explaining violated execution invariants. Appendix~\ref{app:bank_evolution} provides the complete recorded examples.

\subsection{PI Strategy Utility Depends on Student State}
\label{sec:pi_analysis}
\begin{figure}[!htb]
\centering
\includegraphics[width=\linewidth]{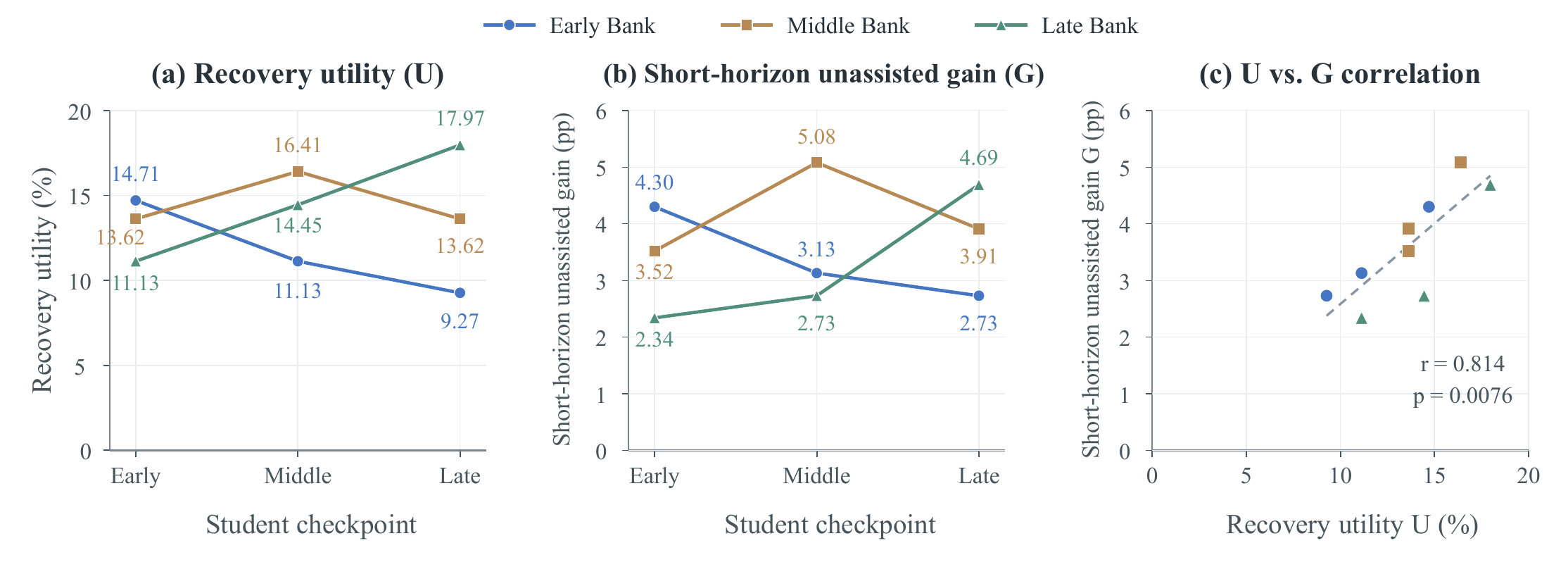}
\caption{\textbf{Student-bank compatibility across training stages.} We compare early-, middle-, and late-stage PI Strategy Banks across three student checkpoints. (a) Recovery utility $U$ for each bank-student pairing. (b) Short-horizon unassisted gain $G$ after training from each checkpoint with the corresponding bank. (c) Association between $U$ and $G$ across all nine pairings. The stage-matched bank achieves the highest $U$ and $G$ at each checkpoint.}
\label{fig:compatibility}
\end{figure}

To test whether an acquisition strategy is intrinsically better or instead matched to a particular student state, Figure~\ref{fig:compatibility} crosses three bank snapshots with early, middle, and late student checkpoints. Recovery utility $U$ measures assisted repair at a fixed checkpoint, while short-horizon gain $G$ measures the resulting change in unassisted performance after a short training branch (Appendix~\ref{app:paired_diagnostics}).

The corresponding-stage bank performs best under both metrics at every checkpoint. In particular, the relative ordering of the early and late banks reverses as the student changes, while the middle bank is preferred by the middle student. Strategy evolution therefore does not produce a universally better sequence of banks; its utility depends on the student that consumes the feedback.

Across the nine student--bank pairs, $U$ and $G$ are strongly associated (Pearson $r=0.814$, two-sided $p=0.0076$). Their rankings are not identical, and the pairs share student checkpoints, but the observed agreement supports recovery utility as a useful selection signal for adapting PI acquisition to the current student and its changing feedback needs.

\section{Related Work}
\label{sec:related}
\paragraph{Knowledge Distillation.}
Knowledge distillation transfers knowledge from a stronger teacher to a smaller student through supervision beyond the original training labels. In generative tasks, sequence-level distillation trains students on teacher-generated responses~\citep{kim2016sequence}, offering a simple way to transfer generation behavior but introducing a mismatch between teacher-generated trajectories and the states actually visited by the student. This mismatch has motivated methods that construct supervision closer to the student's own generation process. The challenge becomes more pronounced for proprietary teachers, whose logits, hidden states, and parameters are inaccessible and whose knowledge must therefore be transferred through observable outputs. EvoOPD studies this black-box setting and focuses on how teacher information can be acquired and converted into supervision on student-generated trajectories.

\paragraph{On-Policy Distillation.}
On-policy distillation addresses trajectory mismatch by constructing supervision directly on student responses~\citep{agarwal2024policy}. Recent work has explored denser and more reliable supervision through privileged contexts, sample routing, tokenizer alignment, reliability modeling, and improved sampled-token objectives~\citep{zhao2026self,zhang2026latent,yang2026self,li2026unifying,sun2026simct,yang2026influence}. Related analyses further show that teacher continuations may become unreliable at student-induced states and that useful supervision can concentrate on a small subset of context-sensitive tokens, motivating selective control over where and when teacher signals are applied~\citep{zhu2026reorder,liu2026visual}.

Most dense OPD methods, however, still assume access to teacher probabilities or another directly comparable supervision signal. For black-box teachers, textual interaction provides an alternative interface. ROPD uses textual rubrics to supervise student responses~\citep{fang2026rubric}, while curriculum turn-level guidance introduces teacher-generated interventions during training, although its distillation objective still relies on teacher distributions~\citep{li2026guidedopd}. More broadly, ProTeGi, EvoPrompt, and Promptbreeder optimize textual instructions, while REvolve and EvoLM adapt reward functions or evaluation rubrics from learning experience~\citep{pryzant2023automatic,guo2024connecting,fernando2023promptbreeder,hazra2025revolve,li2026evolm}. EvoOPD is related to these adaptive-supervision approaches, but optimizes a different object: the strategy for acquiring privileged information from a black-box teacher, including what information to request, where to request it, and at what granularity, with these choices periodically reassessed as the student evolves. This distinction separates adaptation of the information-acquisition policy from adaptation of the student-facing supervision objective: a bank revision changes which evidence is requested while the learning objective remains fixed. Evaluating these revisions on the current student also allows a reusable acquisition rule to be reconsidered even when it continues to produce different feedback for each new response.

\section{Conclusion}
\label{sec:conclusion}
EvoOPD treats black-box on-policy distillation as both a learning problem and an information-acquisition problem. It acquires response-specific textual PI under an evolving Strategy Bank and converts that information into localized supervision on the student's own trajectories. Across mathematics and code with three student scales, EvoOPD consistently improves the final unassisted student over the strongest competing baselines. Ablations, bank-selection dynamics, and student--bank matching further show that the usefulness of a PI acquisition strategy changes with student state, supporting continued adaptation during distillation.

Our current experiments cover one black-box teacher and one student-model family, leaving cross-teacher and cross-family generalization for future study. Recovery utility provides a useful signal for strategy selection but does not perfectly track downstream learning gains, and the local likelihood response to PI does not guarantee semantic correctness. Extending EvoOPD with cross-model transfer, semantic PI validation, and student-adaptive review schedules warrants further study.

% Statements and references follow the main text.
\clearpage
\section*{AI Use Statement}
OpenAI language models accessed through ChatGPT and Codex assisted with literature retrieval and related-work discovery, research brainstorming and feedback on experimental design, interpretation of author-provided experimental results, drafting and editing portions of the manuscript, and translation. The authors reviewed and verified the AI-assisted text, citations, and reported results and take responsibility for the final content of the paper. Language-model calls that constitute part of the EvoOPD research method, including black-box teacher interactions, are described separately in Section~\ref{sec:method} and the appendices.
\section*{Reproducibility Statement}
Section~\ref{sec:method} and Algorithms~\ref{alg:evopi}--\ref{alg:evolve} specify PI acquisition, binary group-relative advantages, local PI correction, and strategy selection. Tables~\ref{tab:math}--\ref{tab:controls} report the supplied main and ablation results; aggregate comparisons use question-count weighting. Appendices~\ref{app:procedure}--\ref{app:bank_evolution} document the protocol, metrics, diagnostics, and qualitative cases, and Appendix~\ref{app:prompts} provides the available prompt templates. Training follows the method-specific configurations specified in Appendix~\ref{app:recipe}; formal runs use seed 42.
\bibliography{references}
\bibliographystyle{paper_references}
\clearpage
\appendix
\section{Complete Training and PI Strategy Bank Update Procedures}
\label{app:procedure}
Algorithm~\ref{alg:evopi} gives the complete online training loop, including the data-step review schedule and state commits. Algorithm~\ref{alg:evolve} is shared by pre-training bank optimization and continued bank updates. It changes only the bank proposal and selection path. The binary verifier, PI-corrected student objective, and evaluation protocol are held fixed.

\begin{algorithm}[htbp]
\caption{\method: online binary GRPO with evolving PI acquisition}
\label{alg:evopi}
\begin{algorithmic}[1]
\Require Initial student $\theta_0$, seed bank $\bank_{\rm init}$, frozen data manifests
\State Set $\theta\gets\theta_0$; freeze reference $\theta_{\rm ref}\gets\theta_0$
\State Set $B\gets\lceil Q_{\rm train}/32\rceil$; $b_j\gets\lceil jB/9\rceil$ for $j=1,\ldots,9$
\State Gather 64-question initial experience; optimize bank on $\data_0^{\rm sel}$ to obtain $\bank_0$
\State Set $\bank\gets\bank_0$; evaluate $\theta_0$ on $\data_{\rm val}$
\For{data-step index $b=1,\ldots,B$}
\State Take next up to 32 training questions; generate eight fresh unassisted responses each
\State Verify binary outcomes and acquire PI for current errors under $\bank$
\State Keep all original responses; resolve PI windows and record behavior log-probabilities
\State Compute group-relative $A$ by Equation~\ref{eq:binary_advantage}
\State Fix pre-update old-policy log-probabilities and initial-model reference log-probabilities
\For{each consecutive mini-batch of up to eight questions, without shuffling}
\State Score original tokens with and without PI; detach $\delta$ and $\widetilde A$ by Equations~\ref{eq:pi_delta}--\ref{eq:semantic_gate}
\State Update $\theta$ once using Equation~\ref{eq:loss}; record the optimizer update
\EndFor
\State Commit trajectories, PI, optimizer and RNG states, and data cursor; synchronize updated $\theta$ for the next rollout
\If{$b=b_j$ for some $j\in\{1,\ldots,9\}$}
\State Save student state; evaluate on $\data_{\rm val}$
\If{$j<9$}
\State Freeze interval-end $\theta$; select 64 interval questions by a fixed deterministic rule
\State $\bank\gets\operatorname{UpdateBank}(\bank,E,\theta,\data_j^{\rm sel})$ via Algorithm~\ref{alg:evolve}
\EndIf
\EndIf
\EndFor
\State \Return student with best validation accuracy; ties prefer earlier checkpoint
\end{algorithmic}
\end{algorithm}

\begin{algorithm}[t]
\caption{PI Strategy Bank update through one proposal and independent comparison}
\label{alg:evolve}
\begin{algorithmic}[1]
\Require Current PI Strategy Bank $\bank$, 64-question experience $E$, frozen student $\theta^\star$, selection manifest
\State Build compact proposal packet from $E$ only, retaining eight originals per question
\State Request one candidate $\widetilde{\bank}$ with hypothesis, uncertainty, and edit
\If{candidate fails schema, edit, or policy validation}
\State Save INVALID and its reason; \Return $\bank$
\EndIf
\State Freeze $\widetilde{\bank}$; access this round's prefrozen disjoint 128-question manifest
\State Generate 8 shared original responses per selected question, once
\State Verify originals; let $\mathcal E$ contain every original error
\If{$|\mathcal E|=0$}
\State Save no-error decision; \Return $\bank$
\EndIf
\For{$B$ in $\{\bank,\widetilde{\bank}\}$}
\State Diagnose the shared responses under $B$ in independent API contexts
\For{each original error $i$}
\If{$B$ supplies usable PI and a resolved anchor}
\State Continue from $B$'s own original prefix using $\theta^\star$ and paired seed $\xi_i$
\State Verify the final answer; persist output and outcome
\Else
\State Record zero repair and the acquisition failure category
\EndIf
\EndFor
\State Compute $U(B)$ with the common denominator $|\mathcal E|$
\EndFor
\State Block on infrastructure failure; never convert it to zero repair
\State Report repair counts, usable PI, failures, and $\Delta U$
\State Atomically save accept iff $U(\widetilde{\bank})>U(\bank)$, otherwise retain current bank
\State \Return selected bank; restore saved decisions without repeated calls or submission
\end{algorithmic}
\end{algorithm}

\paragraph{Proposal experience.}
The proposal protocol selects 64 question groups from the just-completed data interval by a prespecified deterministic rule and retains all eight original responses in rollout-index order. Pre-training bank optimization uses the separate 64-question $\data_{\mathrm{pre}}$ at the initial student. Selection for the packet is independent of progress scores, gate coverage, and PI outcome. Paired continuations use the frozen interval-end student; original trajectories can come from earlier checkpoints in that interval. Both identities are recorded. Questions used for continued proposal experience remain part of $\data_{\mathrm{train}}$, not an additional held-out allocation.

An eligible original error has format-accepted PI and a resolved target window. Paired continuations with and without PI preserve the original prefix token IDs and share a checkpoint and seed. Each arm uses temperature 1, top-$p$ 1, and at most 8192 new tokens, with thinking disabled. PI is system guidance. Context overflow blocks without truncating the prefix. The four correctness categories describe noisy PI outcomes, not deterministic diagnoses of why a rule succeeds or fails.

For an eligible original error, let $h_i$ be the original prefix before the target window. With the frozen interval-end student $\theta^\star$ and shared seed $\xi_i$, the paired continuations are
\begin{equation}
\begin{aligned}
\hat y_i^{\,0}&=h_i\oplus\operatorname{Sample}(S_{\theta^\star};x,h_i,\xi_i),\\
\hat y_i^{\,z}&=h_i\oplus\operatorname{Sample}(S_{\theta^\star};x,h_i,z_i,\xi_i).
\end{aligned}
\label{eq:paired_feedback}
\end{equation}
Both arms preserve the same original prefix and decoding settings; only the PI context differs.

\paragraph{Compact input.}
The packet contains the current bank, interval summary, rule statistics, and selected examples. A correct original retains its index, correctness, conventional progress score one, and empty PI. For an incorrect response with successful localization, the packet retains the complete located paragraph. Otherwise it uses the last nonempty paragraph labeled \texttt{last\_paragraph\_fallback}; this is not an established error location. A single-paragraph response may therefore be retained in full. No paragraph, question, PI, or diagnostic explanation is silently truncated.

The packet preserves progress evidence, PI, format rejection, abstention reason, and overlap diagnostics. Metadata exposes only rule IDs and anchor status. Localization success does not independently confirm the mathematics or rule out later self-correction. A PI rejected for another reason can still have a resolved anchor. Paired records add each arm's first complete continuation paragraph, final correctness, and finish reason. If the first paragraph remains incomplete at the generation limit, its excerpt is null. Full original and continued responses remain in audit records.

Statistics cover only selected questions. Valid supplied rule IDs determine attribution; missing IDs remain unattributed. Multi-rule records may contribute to several rule counts, so those counts are not a partition. Format acceptance is never relabeled semantic validity. The \texttt{response\_content\_copied} flag remains diagnostic rather than an additional rejection rule.

Input construction uses a whitelist. It excludes gold answers, hidden tests, probabilities, likelihood gaps, gates, KL, short-horizon unassisted gains, and final-test outcomes. Stable compact JSON is checked against 800000 local input tokens; the output validation ceiling is 16384. Excess input blocks with its evidence intact rather than removing questions or adding a truncation fallback.

\paragraph{Candidate interface and validation.}
Gemini returns \texttt{hypothesis}, \texttt{uncertainty}, \texttt{edit}, and the complete \texttt{bank}. It identifies a recurring information need, explains a plausible bank-addressable mismatch, and records alternative explanations. It is asked to propose one minimal reusable candidate, preferring REFINE over ADD when possible. It does not output a final KEEP or UPDATE decision.

The configured bank size is 1--8 rules, with one bounded ADD, REFINE, SPLIT, MERGE, or REMOVE edit. Each edit cites at least two distinct question IDs in the current proposal packet; repeated rollouts of one question count once. The program checks membership and count, not whether those questions semantically prove the hypothesis. These numerical limits are implementation choices rather than established optimal settings.

Only \texttt{condition}, \texttt{knowledge\_request}, \texttt{target\_policy}, \texttt{granularity}, and \texttt{combination\_guidance} may change. Existing rule IDs are preserved. IDs required by additions or splits or merges are assigned deterministically by the program; untouched rules cannot be renamed. Field, topology, size, and disclosure checks apply to the complete bank. An unchanged, malformed, or out-of-scope candidate is INVALID, retains the current bank, and does not trigger another proposal.

\paragraph{Selection-set construction.}
The protocol fixes all nine disjoint 128-question bank-selection sets before training, following deduplication by question ID, normalized text, and prespecified near-duplicate checks. These sets exclude other held-out roles, gradient questions, and external tests. Round $k\in\{0,\ldots,8\}$ accesses only its designated $\data_k^{\mathrm{sel}}$, after the candidate has been frozen. Questions are not changed based on difficulty or candidate performance; disjointness is required for every comparison.

The frozen current student generates eight shared original responses per question once. Current and candidate banks diagnose these records in independent API contexts and continue from their own resolved original prefixes. The checkpoint, paired seed, temperature 1, top-$p$ 1, and 8192-new-token limit are common. There is no selection-stage no-PI continuation. All original errors form the denominator, including abstentions, rejected PI, and localization failures. Infrastructure failure blocks comparison. A candidate is accepted only when its repair count is strictly greater; ties retain the current bank. With no original errors, utility is undefined and the current bank is retained without resampling. The review record specifies $|\mathcal E|$, both repair counts, usable-PI counts, acquisition failures, and $\Delta U$. Invalid, rejected, or no-error reviews consume one opportunity without another candidate.

\paragraph{Isolation and recovery.}
Request and generation identities include protocol revision, data manifest, review index, bank, student checkpoint, shared original response, each bank's prefix and PI, paired seed, and decoding parameters. Old PI formats and updater receipts cannot substitute for a new request. The recovery protocol saves responses before grading and reuses identity-matched caches and receipts. Structural repair and transport recovery are accounted for separately from new logical proposals. Independent API diagnoses are not assumed deterministic when the provider does not support a fixed seed.

After comparison, the current and candidate bank versions and the selection decision are recorded. Invalid candidates and non-winning comparisons never prompt a search for another edit. Proposal contexts contain only recent experience; selection questions and outcomes are absent from that request. Offline probes record decisions without changing the bank used for training.

\FloatBarrier
\section{Implementation and Data Specification}
\label{app:details}
\subsection{Final training protocol}
\label{app:recipe}
The formal protocol covers Qwen3-1.7B, 4B, and 8B for mathematics and Python code. It uses full fine-tuning over one pass of the frozen $\data_{\mathrm{train}}$, eight responses per question, and up to 32 questions per data step. The last step retains all remaining questions without padding or dropping them. Binary correctness gives $A_i=c_i-\frac18\sum_j c_j$; group standard deviation is not used. CPU AdamW uses learning rate $10^{-6}$, weight decay $0.01$, $(\beta_1,\beta_2)=(0.9,0.999)$, epsilon $10^{-8}$, and gradient-norm clipping one, without warmup or a scheduler. Training seed is 42. Rollouts use temperature 1, top-$p$ 1, unrestricted top-$k$, and at most 8192 new tokens. Each complete data step uses four ordered eight-question mini-batches, one PPO epoch, and four optimizer updates; a partial final step uses a smaller last mini-batch. Updated students generate fresh on-policy trajectories for the next step. Correct and PI-unavailable responses remain in the base GRPO objective. A zero-advantage group is not assumed to imply a zero optimizer update because the KL regularizer and optimizer state remain active.

The hyperparameters in Equations~\ref{eq:semantic_gate}--\ref{eq:loss} are $\tau=0.02$, $\lambda=0.5$, $\kappa=3$, and $\beta=0.001$. PI-conditioned scoring uses the same current student with no gradient. The difference between its PI and plain log-probabilities on an original token is detached. A valid window token receives the correction in Equation~\ref{eq:semantic_gate} only when $A_i<0$ and $\delta_{i,t}<-0.02$; clipping $\delta$ to $[-3,3]$ bounds the additional negative advantage by $0.5|A_i|$. The ordinary actor pass carries gradients. PPO uses lower and upper ratio clip margins of $0.20$ and $0.28$, respectively (ratio interval $[0.80,1.28]$), negative dual clip $3$, token-mean loss, entropy coefficient zero, and token-level rollout importance correction capped at $2$ without batch normalization. The old-policy log-probabilities are fixed before the data step's mini-batch updates; actual behavior log-probabilities are retained separately for importance correction. A separate low-variance KL loss with coefficient $0.001$ uses the frozen initial Base model as reference. Neither this reference nor the PI scoring pass supplies a full-vocabulary distillation target. Checkpoints bind student, optimizer, RNG, trajectories, rewards, PI, and data cursor so committed updates are never repeated.

For $B=\lceil Q_{\mathrm{train}}/32\rceil$, the data-batch boundaries $b_j=\lceil jB/9\rceil$, $j=1,\ldots,9$, are fixed before training. Pre-training bank optimization precedes the first batch; continued review occurs at $b_1,\ldots,b_8$. Actual optimizer updates are logged separately and do not trigger bank updates. The checkpoint-selection protocol evaluates initialization and all nine segment endpoints using greedy decoding, one sample per question, and at most 8192 new tokens. Best validation accuracy selects the final checkpoint, with earlier checkpoints winning ties. Identical-weight validation results are reused while retaining all ten scheduled exposure records. Baselines use the same validation set and exposure fractions even when their batch sizes differ. External tests never select checkpoints. There is no separate online per-PI semantic reviewer.

\paragraph{Bank-optimization ablations.}
Fixed Strategy Bank retains the seed bank. Pre-optimized Bank performs one pre-training proposal and selection review and then fixes the resulting bank. EvoOPD uses that pre-optimized bank and eight further reviews at the prescribed data-step boundaries. Offline-9 has the same number of bank-optimization opportunities as EvoOPD: nine proposal and selection reviews, all conducted at the initial student checkpoint before gradient training. It then holds the selected bank fixed during student training. These definitions match the number of review opportunities and specify when they occur.

\paragraph{Localization-only control.}
\label{app:localization_control}
Localization-only uses the same binary rewards, eight responses per question, unstandardized advantages, PPO settings, reference-KL term, and bank-review schedule as EvoOPD. It acquires PI to locate error windows but omits the PI-conditioned scoring pass. Let $M_{i,t}=\mathbf1[t\in W_i]\mathbf1[A_i<0]$, and let $c^{\mathrm{gate}}_{i,t}=\widetilde A_{i,t}-A_i$ denote EvoOPD's detached correction in Equation~\ref{eq:semantic_gate}. The control uses
\begin{equation}
\widetilde A^{\mathrm{loc}}_{i,t}
=\operatorname{sg}[A_i-\alpha M_{i,t}|A_i|],\qquad
\alpha=\frac{\sum_{i,t}(-c^{\mathrm{gate}}_{i,t})}
{\sum_{i,t}M_{i,t}|A_i|}.
\label{eq:localization_control}
\end{equation}
The sums for calibration use independent data, giving $\alpha=0.004968595410162241$, which is then fixed throughout training. Calibration matches the total advantage correction on those data; it does not match tokenwise corrections, gradients, or later parameter trajectories. This comparison removes both threshold filtering and likelihood-dependent weighting, so its performance gap cannot be attributed to threshold filtering alone. Localization-only attains 21.35\% in mathematics and 17.68\% on LCB v6, compared with EvoOPD's 24.97\% and 20.21\%, respectively.

\paragraph{Model contexts.}
The plain and PI-conditioned student passes share the model's tokenizer and non-thinking chat-template setting, explicitly disabling thinking. PI is inserted as system guidance only in the no-gradient scoring pass. Only the original question, original trajectory prefix, and visible PI enter that context. Gemini-3.6-Flash uses high thinking for both diagnosis and proposals, in independent request contexts. Micro-batching must preserve the effective mini-batch, original-token alignment, and four optimizer updates per complete data step.

\paragraph{Baseline recipes.}
All trained baselines use the common $\data_{\mathrm{train}}$ question pool and exclude the reserved roles. SeqKD learns verified teacher targets on that pool. Available settings specify mathematics batch 32 for 4B and 128 for 8B at learning rate $10^{-6}$, and 1.7B code learning rate $5\times10^{-7}$. 

GRPO uses full fine-tuning, eight rollouts per question, mini-batches of eight questions, one PPO epoch, and learning rate $10^{-6}$. The standalone mathematics GRPO baseline assigns rewards of $+1$ to correct responses and $-1$ to incorrect responses, standardizes group advantages by their standard deviation, and uses symmetric clip margins of $0.20$. It has neither an actor-KL nor a reward-KL term and disables additional rollout importance correction. In contrast, the No PI ablation retains EvoOPD's binary rewards, unstandardized advantages, asymmetric clipping, separate reference-KL term, and capped token-level rollout importance correction. No PI therefore isolates the added PI correction within EvoOPD's training recipe; it is not a rerun of standalone GRPO. On-policy ROPD uses full fine-tuning, batch 128 questions, eight rollouts, mini-batch eight, learning rate $10^{-6}$, and reference KL coefficient 0.001.

OPSD uses batch 128, a frozen teacher, non-thinking student and teacher contexts, and full-vocabulary forward KL, $D_{\mathrm{KL}}(p\|s)=\sum_v p(v)\log[p(v)/s(v)]$. Recorded learning rates are $10^{-6}$ for code and $5\times10^{-6}$ for mathematics with the 4B and 8B models. Following the official OPSD recipe, LoRA uses rank 64 and alpha 128 on \texttt{q\_proj}, \texttt{k\_proj}, \texttt{v\_proj}, \texttt{o\_proj}, \texttt{gate\_proj}, \texttt{up\_proj}, and \texttt{down\_proj}. Here full-vocabulary refers to the forward-KL distillation target over the vocabulary; it does not mean full-parameter fine-tuning. 

\subsection{Rule and PI serialization}
\label{app:serialization}
\begin{table}[H]
\caption{\textbf{PI Strategy Bank rule schema.} Field names are preserved in the revised proposal. Examples describe strategy content; global output restrictions are enforced separately.}
\label{tab:schema}
\centering\small
\begin{tabularx}{\linewidth}{@{}p{0.29\linewidth}X@{}}
\toprule
Field & Meaning and example content\\
\midrule
\texttt{id} & Unique reusable rule identifier, e.g., constraint formalization.\\
\texttt{condition} & Applicability trigger, e.g., a missing mathematical condition.\\
\texttt{knowledge\_request} & Information to supply, e.g., the missing relation and its interpretation.\\
\texttt{target\_policy} & Which unresolved error to localize; locations remain internal metadata.\\
\texttt{granularity} & Rule-specific detail: a condition cue, causal explanation, minimal action, representation change, or self-check.\\
\texttt{combination\_guidance} & How to combine this guidance with complementary rules.\\
\bottomrule
\end{tabularx}
\end{table}

The bank contains exactly the identifier and five semantic fields in Table~\ref{tab:schema}, each a nonempty string. Mathematics and code use independent five-rule seeds. Global PI restrictions and scoring instructions remain outside the PI Strategy Bank. Mathematics seed IDs are \path{constraint_and_precondition_cue}, \path{local_causal_explanation}, \path{next_step_guidance}, \path{representation_recovery}, and \path{verification_and_coverage_check}. The rules cover constraint and precondition cues, local causal explanations, next-step guidance, representation recovery, and verification or coverage checks. Each retains its supplied granularity. Appendix~\ref{app:prompts} reproduces the mathematics PI-acquisition and candidate-proposal prompts, including progress-score instructions, and the PI-conditioned student-scoring context.

The serialized acquisition interface retains the original rollout group and the complete diagnostic tuple:
\begin{equation}
\{y_i\}_{i=1}^{N}\sim S_{\theta_k}(\cdot\mid x),\qquad
\{(q_i,e_i,m_i,z_i)\}_{i=1}^{N}
=T(x,\{y_i,c_i\}_{i=1}^{N};\bank_k),\quad N=8.
\label{eq:pi}
\end{equation}
Here $z_i$ is visible PI, $m_i$ locates the original error window $W_i$, and $q_i,e_i$ are diagnostic annotations excluded from the loss. Diagnosis is joint, so PI for one response may depend on other responses to the same question. A group with only correct responses requires no black-box diagnosis. Each diagnosis has eight indexed entries with a progress score and explanation, hidden localization metadata, and visible PI as one nonempty string or null. There are no four mandatory headings. Internal metadata contains the paragraph quote, occurrence, rule IDs, and resolved offsets. Only the visible string enters the PI-conditioned scoring context; scores, correctness flags, other responses, and metadata stay outside it. Visible PI is limited to 512 local tokens. Rule instructions are English; generated PI has no required language.

A mathematics step is a nonempty paragraph separated by blank lines, including internal displayed equations. Exact quote and occurrence must resolve to a complete original paragraph. There is no fixed-token substitute or suffix extension. Failure to resolve yields no usable PI.

\paragraph{Code localization contract.}
Code uses a deterministic extraction-and-AST locator with an offset map into the original response. The required sequence is: extract Python from fenced, plain-code, or mixed responses; parse the extracted code; resolve the teacher's anchor to an allowed complete statement or block; map its original character span to student-token offsets; and retain the original token window unchanged. Manual code correction cannot substitute for localization. Extraction, parsing, ambiguous-anchor, and token-mapping failures are recorded separately and retained in the original-error denominators.

The target is an unresolved error whose repair can redirect reasoning. Unavailable PI or unresolved anchors yield $W_i=\emptyset$.

PI must leave the solution to the student. It excludes anchor quotes, erroneous target expressions, future target content, solved intermediate results, final answers, and complete corrected solutions. Overlap alone is a diagnostic, not a blanket rejection. Format and localization checks cannot certify semantic correctness. Retrospective diagnosis retains hindsight even after explicit target copying is removed, and a future-token causality test with PI held fixed cannot detect content already copied into PI.

An incorrect response may abstain if the target cannot be reliably identified. Correct responses have no PI correction but remain in the base GRPO and reference-KL terms; their progress score one is a convention for the verified outcome. Defective progress scores remain unavailable. A scoring-only defect does not reject otherwise usable PI, since the score is absent from the loss. Malformed group output has at most one structured repair; exhausted invalid output retains all records and denominators with no usable PI correction. Numerical, data, and infrastructure failures block with evidence.

\subsection{Data roles and external evaluation}
The sources are DAPO-Math-17K and the OpenThoughts-114k code subset. Each task reserves the following roles once, shared across its prespecified model and method comparisons:
\begin{center}
\small
\begin{tabularx}{\linewidth}{@{}l r X@{}}
\toprule
Role & Questions & Use \\
\midrule
$\data_{\mathrm{val}}$ & 335 math; 275 code & Student checkpoint selection \\
$\data_{\mathrm{pre}}$ & 64 & Initial proposal experience \\
$\data_k^{\mathrm{sel}}$, $k=0,\ldots,8$ & $9\times128$ & Independent bank selection rounds \\
$\data_{\mathrm{diag}}$ & 128 & Shared originals for $U$ and $C$ diagnostics \\
$\data_{\mathrm{micro}}$ & 256 & Eight data batches for each G branch \\
$\data_{\mathrm{probe}}$ & 256 & Independent unassisted G evaluation \\
$\data_{\mathrm{train}}$ & Remaining eligible pool & Gradient training and continued proposal experience \\
\bottomrule
\end{tabularx}
\end{center}
If $Q_{\mathrm{pool}}$ already excludes validation, test overlap, and ineligible questions, then $Q_{\mathrm{train}}=Q_{\mathrm{pool}}-1856$; validation is not subtracted twice. The code validation set is drawn from the subset with passing teacher solutions, not uniformly from the full code pool.

The split protocol fixes near-duplicate rules before splitting and checks exact IDs, normalized text, near duplicates, and known code-benchmark source problems against external tests. It records each question's source, split, and order. All reserved roles are mutually disjoint and excluded from every method's gradient pool. Continued proposal questions are selected from the completed training interval. Selection sets are disjoint across rounds. Diagnostic sets may be reused across prespecified banks and checkpoints, with results excluded from main training, bank acceptance, and checkpoint selection.

The final evaluation spans six mathematics and five code benchmarks. LiveCodeBench versions 5 and 6 overlap is recorded rather than treated as independent replication. BigCodeBench uses Hard v0.2.5. Hidden training-side code tests remain verifier-only.

\subsection{Metrics and access accounting}
\label{app:metrics}
\label{sec:efficiency}
For benchmark $b$ with $N_b$ questions and $R_{\mathrm{eval}}$ completions per question,
\begin{equation}
\mathrm{Acc}_b=\frac{100}{N_bR_{\mathrm{eval}}}
\sum_{i=1}^{N_b}\sum_{r=1}^{R_{\mathrm{eval}}}c_{i,r},
\qquad
\mathrm{Overall}=\frac{\sum_bN_b\mathrm{Acc}_b}{\sum_bN_b}.
\label{eq:mathmetric}
\end{equation}
Final decoding uses temperature 1.0, top-$p$ 0.95, unrestricted top-$k$, at most 32768 new tokens, non-thinking, and seed 42. Mathematics uses 16 samples per question for all three student sizes. Its six benchmarks contain 183 questions: 33 in HMMT26 Feb and 30 in each other benchmark. Code uses 16 samples per question and contains 1,032 questions across the five benchmarks.

Math averages all sampled completions, not best-of-$N$ correctness. Code pass@1 is estimated by the mean correctness of 16 samples per question. Its Overall uses the same question-count weighting as Equation~\ref{eq:mathmetric}, with benchmark sizes 164, 378, 148, 167, and 175; LiveCodeBench v6 is the representative-model ablation outcome. Undefined denominators are unavailable rather than zero.

The accounting protocol separates training PI, pre-training bank optimization and continued bank updates experience, candidate requests, bank-selection calls, and diagnostics. Its fields distinguish logical calls, transport attempts, repairs, retries, input tokens, output tokens, cached tokens, reasoning tokens, and provider charges. Local generation, scoring, optimization, and evaluation are separate compute categories.

\paragraph{Reported teacher-token usage.}
\label{app:api_budget}
Table~\ref{tab:api_budget} in Section~\ref{sec:ablation} reports the available mathematics totals. Input, output, and thinking tokens are separate categories in the supplied records, and their sums give the displayed totals. No PI makes no teacher calls. The Fixed Strategy Bank token total is unavailable and remains a dash rather than zero. EvoOPD uses 2.84 million more tokens than Offline-9, or 1.54\%, while improving mathematics Overall by 1.16 percentage points. These are similar recorded token totals, not a controlled equal-budget comparison. They do not establish equal monetary cost, wall-clock time, or local compute requirements.

\section{Mechanism Diagnostics and Failure Analysis}
\label{app:analysis}
\subsection{Student state and PI utility}
\label{app:paired_diagnostics}
Figure~\ref{fig:compatibility} crosses three prespecified bank snapshots $\bank_i$ with fixed student checkpoints $\theta_j$ on $\data_{\mathrm{diag}}$. Early, middle, and late correspond to S1, S4, and S8, respectively. The middle snapshot retains the S3 bank because S4 rejects a tied candidate. Within checkpoint $j$, all banks share original unassisted responses and paired seeds while selecting their own anchors and PI. The same set $\mathcal E_j$ of original errors determines all recovery denominators.

\paragraph{Recovery utility.}
Define $U_{ij}=U(\bank_i;\theta_j,\data_{\mathrm{diag}})$ using Equation~\ref{eq:selection_utility}: successfully repaired originals divided by all original errors. This diagnostic uses independent questions and does not reuse the selection sample. An error without usable PI contributes zero; infrastructure failure blocks. The contrasts between continuations with and without PI used in proposal experience remain a different measurement. Undefined denominators are reported as unavailable. Figure~\ref{fig:compatibility}(a) organizes recovery by student checkpoint and bank snapshot.

\paragraph{Short-horizon unassisted gain.}
The diagnostic protocol clones the student, optimizer, and RNG state at $\theta_j$ into one branch per bank. Each branch uses the same 256 $\data_{\mathrm{micro}}$ questions in $J=8$ batches of 32, eight fresh on-policy trajectories per question, and its own fixed bank under the unchanged loss, obtaining $\theta'_{j,i}$. The short-horizon unassisted gain is
\begin{equation}
G_{ij}=\mathrm{Perf}_{\data_{\mathrm{probe}}}(\theta'_{j,i})
-\mathrm{Perf}_{\data_{\mathrm{probe}}}(\theta_j).
\label{eq:proxy}
\end{equation}
Evaluation uses 256 independent $\data_{\mathrm{probe}}$ questions, greedy decoding, one sample per question, and at most 8192 new tokens, identically before and after each branch. Branches share the initial optimizer and RNG states, fixed Base reference, question order, and sampling recipe. Eight complete data steps entail 32 mini-batch optimizer updates; an empty PI gate removes only the added correction, not the base GRPO and KL terms. Figure~\ref{fig:compatibility} reports the cross-checkpoint comparison.

\paragraph{Gate coverage.}
For bank-specific original windows and token gates from Equation~\ref{eq:semantic_gate}, define trajectory-level PI coverage as
\begin{equation}
C_{ij}=\frac{1}{|\mathcal E_j|}\sum_{r\in\mathcal E_j}
\mathbf1[\exists t:\ g_{r,t,\bank_i}=1].
\label{eq:gate_coverage}
\end{equation}
PI-unavailable and unlocalized errors remain in this denominator. Coverage measures availability of an additional PI advantage correction, not admission to base GRPO, and is distinct from candidate-bank acceptance. Figure~\ref{fig:compatibility}(c) instead plots the association between $U$ and $G$. Token-level coverage is a supplementary diagnostic.

\paragraph{Interpretation and supplemental curves.}
Figure~\ref{fig:compatibility} shows changing bank rankings across student checkpoints. The corresponding-stage banks achieve the highest $U$ (14.71\%, 16.41\%, and 17.97\%) and $G$ (4.30, 5.08, and 4.69 percentage points) for the early, middle, and late students, respectively. Panel (c) uses all nine corresponding pairs and an ordinary least-squares line, giving Pearson $r=0.814$ and two-sided $p=0.0076$. The nine observations share student checkpoints and are aggregate comparisons, not independent training seeds. Figure~\ref{fig:continued}(a) reports six accepted revisions, one negative difference at S2, and a tie at S4; both nonpositive candidates are rejected. Panel (b) compares LiveCodeBench v6 outcomes, starting at 15.39\% and ending at 17.84\% for Fixed Strategy Bank, 18.92\% for Pre-optimized Bank, and 20.21\% for EvoOPD.

\paragraph{Binary advantage and local PI effect.}
Progress scores describe errors for bank refinement but do not determine rewards or enter the policy loss. For $m$ correct responses among eight, every correct response has $A=1-m/8$ and every incorrect response has $A=-m/8$. All-correct and all-wrong groups have $A=0$ and therefore no PI correction. In mixed groups, an error token is PI-eligible only when the same student assigns it a PI-conditioned log-probability at least $0.02$ below the plain-context value. The correction in Equation~\ref{eq:semantic_gate} is detached, nonpositive, and bounded; it cannot turn an incorrect response into a positive-reward example. This likelihood test does not certify that PI is semantically correct.

For a valid error window, the aggregate diagnostic
\begin{equation}
\Gamma_i=\sum_{t\in W_i}\delta_{i,t}
\label{eq:pi_effect}
\end{equation}
holds student weights, original tokens, and prefixes fixed while changing only the PI context. Negative $\Gamma_i$ can reflect style or probability redistribution and does not by itself establish unassisted learning. Trajectory coverage can be normalized by all original errors; accepted-token coverage uses candidate-window tokens as its denominator. Undefined denominators remain unavailable.

\paragraph{Scope of recovery utility.}
Recovery utility measures the combined effects of localization, PI availability, and correction success at a fixed student checkpoint. The agreement between $U$ and $G$ concerns the best bank at each checkpoint and their positive association across the nine pairs, rather than identical full rankings. Longer-horizon unassisted performance can also depend on optimization dynamics and supervision coverage. The No PI condition in Table~\ref{tab:controls} is a separate ablation control, with 16.46\% mathematics Overall and 15.18\% LiveCodeBench v6 pass@1; it is not the untrained Base row in Tables~\ref{tab:math} and~\ref{tab:code}. The ablation conditions use their own on-policy responses; shared data exposure does not require identical realized trajectories or target windows.

\paragraph{Signal--learning alignment.}
Recovery utility $U$ and short-horizon unassisted gain $G$ both rank the early bank above the middle and late banks at the early checkpoint, and rank the late bank above the middle and early banks at the late checkpoint. At the middle checkpoint, both metrics prefer the middle bank, but the late bank has higher $U$ than the early bank (14.45\% versus 11.13\%) and lower $G$ (2.73 versus 3.13 points). Thus repair utility does not perfectly order all downstream gains. $\Gamma$, accepted-token coverage, and progress summaries characterize PI availability; bank acceptance uses only Equation~\ref{eq:selection_utility}.

\paragraph{Failure taxonomy.}
\label{app:failure}
\textbf{A: Incorrect diagnosis or PI.} The localized error or supplied constraint is wrong.
\textbf{B: Failed local use.} Valid PI does not yield useful supervision.
\textbf{C: Assistance--learning mismatch.} Recovery utility improves without short-horizon unassisted gain.
\textbf{D: Refinement regression.} A candidate wins bank-selection recovery but degrades subsequent training.
These categories describe possible failures rather than measured frequencies. The local likelihood gate cannot rule out semantic PI errors or imperfect localization.

\paragraph{Future analyses.}
Positive-preservation objectives, Always Accept, alternative PI thresholds, a score-blind updater, and retrospective human failure audits are unexecuted extensions. They could examine which parts of the acquisition and selection procedure contribute to unassisted performance; cross-teacher and cross-family transfer are further directions.

\section{Prompt Templates}
\label{app:prompts}
The following mathematics prompts are supplied with runtime inputs. Their original term ``PI Rule Bank'' denotes the PI Strategy Bank. Progress scores and evidence support diagnosis and proposals, not the student loss. The framed listings can continue across pages.

\subsection{Joint PI acquisition for eight responses}
\label{app:prompt_acquisition}
The instruction is sent with the current bank, the question, eight original responses, and their correctness flags. The bank is fixed during each diagnosis call. Line wrapping below is typographical.
\tcbinputlisting{casepi,
  title={PI acquisition instruction},listing only,listing engine=listings,
  listing options={style=paperprompt,frame=none},listing file=prompts/pi_acquisition.txt}

\paragraph{Runtime payload and structured output.}
The appended payload has the fields \texttt{current\_bank}, \texttt{question}, and \texttt{eight\_responses}. Each response supplies its index, text, and correctness-verifier label; gold answers and hidden tests are absent. The enforced output contains exactly eight entries with \texttt{rollout\_id}, diagnostic \texttt{progress\_score}, \texttt{score\_evidence}, and \texttt{abstention\_reason}, hidden \texttt{anchor\_metadata}, and \texttt{visible\_pi}. Only \texttt{visible\_pi} enters the no-gradient PI-conditioned student context. Localization metadata, scores, explanations, correctness flags, and other responses stay outside it.

\subsection{One bounded PI Strategy Bank candidate}
\label{app:prompt_proposal}
The candidate instruction is followed by the compact experience packet: current bank, recent questions, original error windows, PI, progress evidence, paired continuation outcomes, and program-allocated rule IDs. The separate bank-selection sample is unavailable to the proposer.
\tcbinputlisting{caserepair,
  title={Bank proposal instruction},listing only,listing engine=listings,
  listing options={style=paperprompt,frame=none},listing file=prompts/bank_proposal.txt}

\subsection{PI-conditioned student scoring}
\label{app:prompt_teacher}
For valid original error windows, the same current student scores the unchanged original target tokens in two contexts. The PI branch has no gradient; it receives visible PI as system guidance followed by the original question and original response-prefix token IDs. The plain branch receives the original question and identical prefix without PI. No corrected answer is generated for the actor update. The token log-probability difference is detached before the gated advantage correction; neither PI tokens nor future target tokens are supervised targets.

\section{Qualitative Mathematics Cases}
\label{app:cases}
These two Qwen3-1.7B examples illustrate how a localized PI cue can change a student's continuation. They are selected success cases, not an estimate of PI success frequency. The original prefix before the identified error is retained; only the suffix is regenerated after PI is supplied as system guidance. The excerpts show the error transition rather than the complete trajectories. B0 and B8 identify the initial and an evolved PI bank.

\subsection{Case 1: Restoring an omitted probability region (B0)}
Alice and Bob arrive independently and uniformly between 1:00 and 2:00. Given that Alice arrives after Bob, the question asks for the probability that Bob arrived before 1:30, expressed as $m/n$, and then $m+n$. Let $X,Y\in[0,60]$ be Alice's and Bob's arrival times in minutes after 1:00.

\begin{tcolorbox}[caseerror,title={Original error}]
The student set up $P(Y<30\mid X>Y)$ correctly but restricted the numerator to $0\leq Y<30$ and $30<X<60$. This omitted the valid region $Y<X<30$ and gave numerator area $900$ instead of $1350$, leading to $900/1800=1/2$ and the wrong answer $m+n=3$.
\end{tcolorbox}
\begin{tcolorbox}[casepi,title={Privileged information}]
When Bob arrives before the midpoint of the hour, Alice can arrive at any time after him, including before the midpoint. Restricting Alice's arrival to only after the midpoint omits the region where both arrive in the first half of the hour. Instead, Alice's arrival time should range from Bob's arrival time to the end of the hour.
\end{tcolorbox}
\begin{tcolorbox}[caserepair,title={Continuation after PI}]
The student changed the bounds to $0\leq Y<30$ and $Y<X<60$. It then used $\int_0^{30}(60-Y)\,dY=1350$ and obtained $P(Y<30\mid X>Y)=1350/1800=3/4$, so the final answer changed from $3$ to $7$.
\end{tcolorbox}

\subsection{Case 2: Restoring a repeated-game constraint (B8)}
A conference has 14 teams in two divisions of seven. Every pair within a division plays twice; every cross-division pair plays once. The question asks for the total number of games.

\begin{tcolorbox}[caseerror,title={Original error}]
The student counted $\binom72=21$ pairs per division and multiplied by two divisions to obtain $42$ intra-division games. It missed the requirement that each within-division pair plays \emph{twice}. Adding $7\times7=49$ cross-division games yielded the wrong total $91$.
\end{tcolorbox}
\begin{tcolorbox}[casepi,title={Privileged information}]
Recall that each team must play every other team in its own division twice. Your current calculation only accounts for each pair of teams playing a single game. You need to adjust the number of games within a single division to reflect that every pair plays two matches.
\end{tcolorbox}
\begin{tcolorbox}[caserepair,title={Continuation after PI}]
The student corrected one division to $\binom72\times2=42$ games, then counted both divisions as $84$. With the unchanged $49$ cross-division games, its final answer changed from $91$ to $84+49=133$.
\end{tcolorbox}

\section{Qualitative Code Cases}
\label{app:codecases}
The following two Qwen3-1.7B examples use the initial bank B0 and show different kinds of code repair: satisfying an output contract and restoring an algorithmic invariant. They are selected successful continuations, not estimates of repair frequency. As above, PI is supplied as system guidance at the localized error, and the original prefix is retained. Only the decisive code transition is shown.

\subsection{Case 1: Printing the required strings}
The program must read a lowercase string and print the case-sensitive word \texttt{yes} if all characters are distinct, or \texttt{no} otherwise. The original solution correctly tested uniqueness with \texttt{len(set(s)) == len(s)} but violated the output specification.

\begin{tcolorbox}[caseerror,title={Original error}]
After reading the string, the student used \texttt{print(is\_unique(S))}. Python therefore printed \texttt{True} or \texttt{False} instead of the required \texttt{yes} or \texttt{no}.
\end{tcolorbox}
\begin{tcolorbox}[casepi,title={Privileged information}]
The problem requires outputting the exact string 'yes' or 'no' depending on whether all characters are unique. Currently, this statement outputs a boolean value (True or False), which violates the required output specification contract.
\end{tcolorbox}
\begin{tcolorbox}[caserepair,title={Continuation after PI}]
The student kept the uniqueness test and changed only the output step to \texttt{print("yes" if is\_unique(S) else "no")}. The continuation now prints the specified string for either verifier outcome.
\end{tcolorbox}

\subsection{Case 2: Expanding the reachable-sum interval}
Given sorted positive integers \texttt{nums} and a target $n$, the program must add as few values as possible so every integer in $[1,n]$ is representable as a subset sum. Let $R$ be the largest value such that every integer in $[1,R]$ is already reachable.

\begin{tcolorbox}[caseerror,title={Original error}]
When the next array element could not fill the gap at $R+1$, the student incremented the patch count but updated the bound with \texttt{max\_reach *= 2}. This fails to include the patch value $R+1$; at $R=0$ it makes no progress at all.
\end{tcolorbox}
\begin{tcolorbox}[casepi,title={Privileged information}]
When patching a gap in the range of reachable sums, the smallest integer that cannot currently be formed is \texttt{max\_reach + 1}. Adding this patch value to the available numbers increases the maximum reachable sum by that patch amount. Doubling \texttt{max\_reach} omits this extra unit, leaving the reachable upper bound smaller than expected and preventing progress when \texttt{max\_reach} is zero. Reconsider how adding a patch of value \texttt{max\_reach + 1} should update the maximum reachable sum.
\end{tcolorbox}
\begin{tcolorbox}[caserepair,title={Continuation after PI}]
The student replaced the faulty update with \texttt{patch = max\_reach + 1} followed by \texttt{max\_reach += patch}. Thus the covered interval grows from $[1,R]$ to $[1,2R+1]$; unlike the original update, it also advances from $R=0$.
\end{tcolorbox}

\section{Examples of PI Strategy Bank Evolution}
\label{app:bank_evolution}
The examples below summarize selected rule changes in mathematics and code. Each chain shows the initial bank, its first revision, and the bank retained after eight reviews; the ellipsis omits intermediate reviews. Descriptions are condensed from the recorded rules, with \textbf{bold} marking changes in emphasis. The same five rule identities remain in these snapshots. These examples describe acquisition content, rather than identifying which individual edit caused a performance gain.

\subsection{Mathematics: Making Local Guidance More Explicit}
\noindent
\begin{minipage}[c]{0.30\linewidth}
\begin{tcolorbox}[casepi,title=Bank v0,breakable=false,height=150pt,valign=top,before skip=0pt,after skip=0pt]
\small\raggedright
\textbf{Constraint cue.} Identify a missing or violated condition and explain its consequence for the current inference. Do not provide the corrected derivation.

\textbf{Scope.} Give a concise local cue, leaving the student to determine how to enforce the condition.
\end{tcolorbox}
\end{minipage}\hfill
\begin{minipage}[c]{0.04\linewidth}\centering\small $\rightarrow$\end{minipage}\hfill
\begin{minipage}[c]{0.30\linewidth}
\begin{tcolorbox}[casepi,title=Bank v1,breakable=false,height=150pt,valign=top,before skip=0pt,after skip=0pt]
\small\raggedright
\textbf{Constraint cue.} Add the \textbf{constructive action or valid equation setup} needed to enforce the condition at the current step.

\textbf{Scope.} State the condition, its consequence, and how to satisfy it, while withholding the final expression or answer.
\end{tcolorbox}
\end{minipage}\hfill
\begin{minipage}[c]{0.045\linewidth}\centering\small $\cdots$\\$\rightarrow$\end{minipage}\hfill
\begin{minipage}[c]{0.30\linewidth}
\begin{tcolorbox}[caserepair,title=Bank v8,breakable=false,height=150pt,valign=top,before skip=0pt,after skip=0pt]
\small\raggedright
\textbf{Constraint cue.} Specify the \textbf{exact missing term or bound} and local structural adjustment.

\textbf{Representation.} Give an \textbf{explicit substitution and first equation}.

\textbf{Verification.} Specify the \textbf{domain, bound, or boundary condition} to check. Leave final answers to the student.
\end{tcolorbox}
\end{minipage}

The first revision changes only the constraint rule's requested knowledge and granularity: identifying a condition becomes actionable local guidance. By the final snapshot, representation recovery and verification are also more explicit. This illustrates refinement of what the teacher should supply within existing rule roles. Bank labels here index review stages: two rejected candidates leave the bank unchanged, so Bank v8 is the sixth accepted-update version in the mathematics record.

\subsection{Code: Preserving Structure and Explaining Execution Errors}
\noindent
\begin{minipage}[c]{0.30\linewidth}
\begin{tcolorbox}[casepi,title=Bank v0,breakable=false,height=150pt,valign=top,before skip=0pt,after skip=0pt]
\small\raggedright
\textbf{Representation.} Suggest an alternative data structure, representation, or algorithmic decomposition. Explain why it exposes the needed structure.

\textbf{Scope.} Leave the transformed code and implementation to the student.
\end{tcolorbox}
\end{minipage}\hfill
\begin{minipage}[c]{0.04\linewidth}\centering\small $\rightarrow$\end{minipage}\hfill
\begin{minipage}[c]{0.30\linewidth}
\begin{tcolorbox}[casepi,title=Bank v1,breakable=false,height=150pt,valign=top,before skip=0pt,after skip=0pt]
\small\raggedright
\textbf{Representation.} Explicitly preserve \textbf{entity boundaries, attribute pairings, or global sequence keys} lost in the current representation.

\textbf{Scope.} Explain the structural rationale for one representation change, without a worked patch.
\end{tcolorbox}
\end{minipage}\hfill
\begin{minipage}[c]{0.045\linewidth}\centering\small $\cdots$\\$\rightarrow$\end{minipage}\hfill
\begin{minipage}[c]{0.30\linewidth}
\begin{tcolorbox}[caserepair,title=Bank v8,breakable=false,height=150pt,valign=top,before skip=0pt,after skip=0pt]
\small\raggedright
\textbf{Representation.} Retain the v1 structural guidance.

\textbf{Constraints.} Include \textbf{input layout and execution scope}.

\textbf{Causal explanation.} Contrast \textbf{actual execution with required invariants} for iteration, counters, ordering, and state transitions. Provide no replacement code.
\end{tcolorbox}
\end{minipage}

The first revision refines representation recovery, making the loss of relational structure an explicit target. The final bank retains that revision and further refines the constraint and causal-explanation rules. The latter asks why an operation violates the intended invariant, while leaving repair construction to the student. Thus the chain shows complementary edits across a bank, not successive rewrites of one rule. Next-step guidance and verification remain unchanged in the displayed code snapshots.

\end{document}